\documentclass[letterpaper, 10 pt, conference]{ieeeconf}  

\IEEEoverridecommandlockouts                              

\usepackage{graphicx, multirow} 
\usepackage{mathptmx} 
\usepackage{amsmath} 
\usepackage{dsfont}
\usepackage{amssymb}  
\usepackage{adjustbox}
\usepackage{wasysym}
\usepackage{subcaption}
\usepackage{nicefrac}
\usepackage{xspace}
\usepackage{comment}
\usepackage{booktabs}
\usepackage{arydshln}
\usepackage[table,x11names]{xcolor}
\usepackage{dsfont}
\usepackage{rotating}
\usepackage{pifont}
\usepackage[noadjust]{cite}

\usepackage{fancyhdr}
\usepackage[ruled,vlined]{algorithm2e}
\include{pythonlisting}

\definecolor{ForestGreen}{RGB}{34,139,34}
\definecolor{Blue}{RGB}{34,64,200}
\definecolor{Orange}{RGB}{255,165,0}
\definecolor{lightgray}{RGB}{240,240,240}
\newcommand{\cmark}{\ding{51}}

\makeatletter
\DeclareRobustCommand\onedot{\futurelet\@let@token\@onedot}
\def\@onedot{\ifx\@let@token.\else.\null\fi\xspace}

\def\eg{\emph{e.g}\onedot} 
\def\ie{\emph{i.e}\onedot}

\makeatother

\newcommand{\myparagraph}[1]{\vspace{4pt}\noindent\textbf{#1}}

\newcommand\ourMix{\text{HIMix}}

\usepackage{my_ref}

\title{\LARGE \bf
Hierarchical Instance Mixing across Domains in Aerial Segmentation
}

\author{Edoardo Arnaudo$^{*1,2}$, Antonio Tavera$^{*1}$, Fabrizio Dominici$^2$, Carlo Masone$^3$, Barbara Caputo$^1$ 
\\
$^1${\small Politecnico di Torino, Turin, Italy
}\\
$^2${\small LINKS Foundation, Turin, Italy}
\\
$^3${\small CINI - Consorzio Interuniversitario Nazionale per l'Informatica, Rome, Italy}
\\
$^1${\{\tt\small edoardo.arnaudo, antonio.tavera, barbara.caputo\}@polito.it}
\thanks{$^*$Equal contribution.}
}

\begin{document}

\maketitle
\thispagestyle{plain}
\pagestyle{plain}


\begin{abstract}
We investigate the task of unsupervised domain adaptation in aerial semantic segmentation and discover that the current state-of-the-art algorithms designed for autonomous driving based on domain mixing do not translate well to the aerial setting. This is due to two factors: (i) a large disparity in the extension of the semantic categories, which causes a domain imbalance in the mixed image, and (ii) a weaker structural consistency in aerial scenes than in driving scenes since the same scene might be viewed from different perspectives and there is no well-defined and repeatable structure of the semantic elements in the images.
Our solution to these problems is composed of: (i) a new mixing strategy for aerial segmentation across domains called Hierarchical Instance Mixing (HIMix), which extracts a set of connected components from each semantic mask and mixes them according to a semantic hierarchy and, (ii) a twin-head architecture in which two separate segmentation heads are fed with variations of the same images in a contrastive fashion to produce finer segmentation maps.
We conduct extensive experiments on the LoveDA benchmark, where our solution outperforms the current state-of-the-art.
\end{abstract}
\section{Introduction}
Semantic segmentation aims to predict, for each individual pixel in an image, a semantic category from a predefined set of labels.
Such a fine grained understanding of images finds numerous applications in aerial robotics \cite{demir2018deepglobe,chen2021_lc_mfanet,tong2020_lc_land,ref_beyond_rgb,ref_resunet_a,ref_seg_aerial_nogueira,wildfire_est, deforestation_est,chiu2020agri_multispectral}, where it has achieved remarkable results 
by leveraging deep learning models trained on open datasets with large quantities of labeled images. 
However, these results do not carry over when the models are deployed to operate on images that come from a distribution (\emph{target domain}) different from the data experienced during training (\emph{source domain}).
The difficulty in adapting semantic segmentation models to different data distributions is not only limited to the aerial setting and it is tightly linked to the high cost of generating pixel-level annotations \cite{cityscapes}, which makes it unreasonable to supplement the training dataset with large quantities of labeled images from the target domain.
A recent trend in the state-of-the-art addresses this challenge using domain mixing as an online augmentation to create artificial images with elements from both the source and the target domain, thus encouraging the model to learn domain-agnostic features \cite{Chen2021-semisup,dacs,daformer,zhou2021context}.
In particular, both DACS \cite{dacs} and DAFormer \cite{daformer} rely on ClassMix \cite{olsson2021classmix} to dynamically create a binary mixing mask for a pair of source-target images by randomly selecting half of the classes from their semantic labels (the true label for the source, the predicted pseudo-label for the target). 
Although this mixing strategy yields state-of-the-art results in driving scenes, it is less effective in an aerial context. We conjecture that this is largely caused by two factors: 

\begin{figure}[t]
    \centering
    \includegraphics[width=1.0\columnwidth]{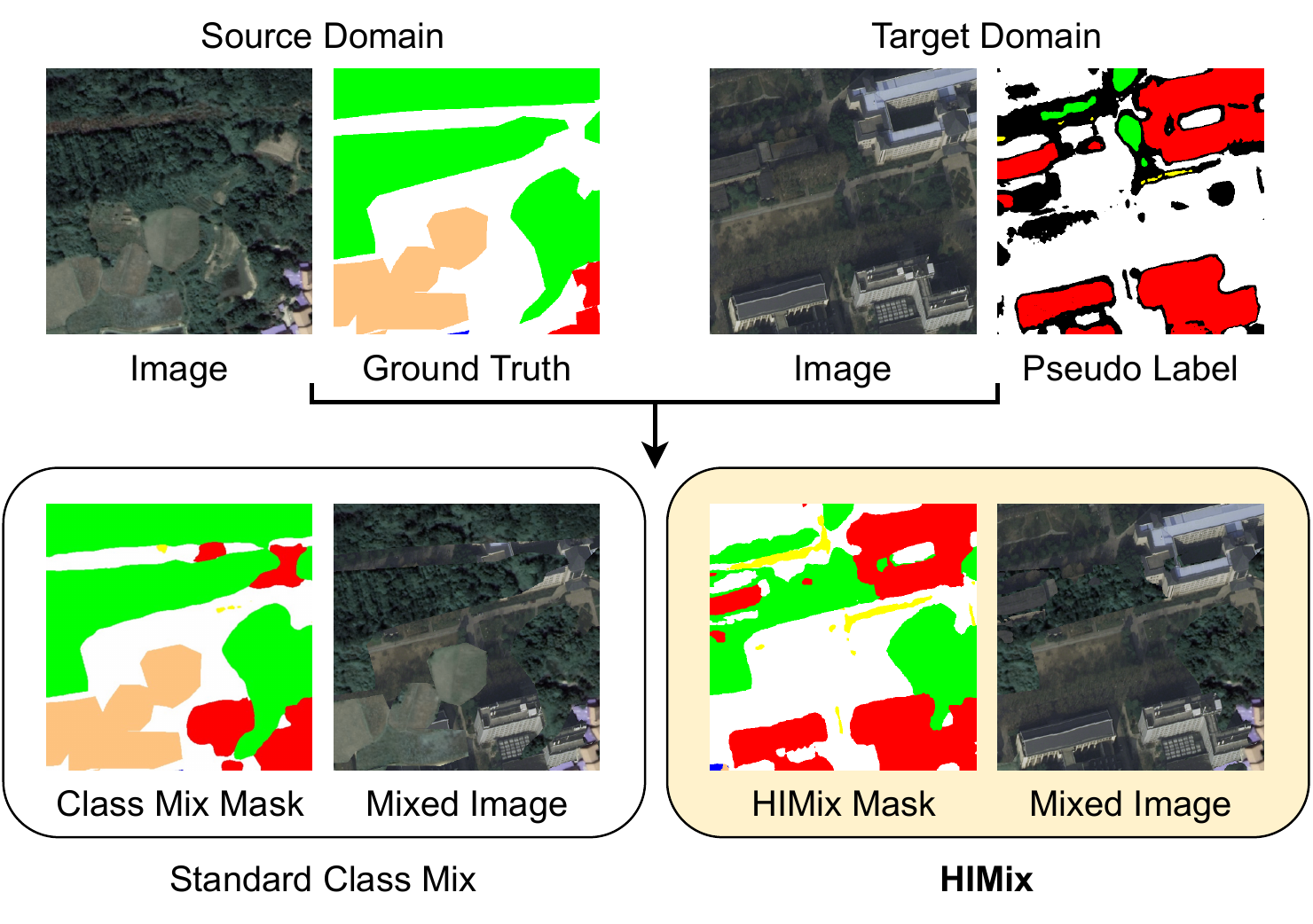}
    \caption{Class Mix superimposes classes of the source domain onto the target without taking into account the semantic hierarchy of the visual elements. As a result, it generates erroneous images that are detrimental to Unsupervised Domain Adaptation training in the aerial scenario. Instead, our \textbf{{\ourMix}} extracts instances from each semantic label and then composes the mixing mask after sorting the extracted instances based on their pixel count. This mitigates some artifacts (e.g. partial buildings) and improves the balance of the two domains.}
    \label{fig:teaser}
\end{figure}

\myparagraph{Domain imbalance in mixed images.}
Segmentation-oriented aerial datasets are often characterized by categories with vastly different extensions (e.g., \textit{cars} and \textit{forest}). While this may be dealt with techniques such as multi-scale training in standard semantic segmentation \cite{ref_multiscale_vhr_seg}, the disparity in raw pixel counts between classes may be detrimental for an effective domain adaptation through class mixing, as the composition may favor either domain (see \cref{fig:teaser} left).

\myparagraph{Weak structural consistency.}
The scenes captured by a front-facing camera onboard a car have a consistent structure, with the street at the bottom, the sky at the top, sidewalks and buildings at the sides, etcetera. This structure is preserved also across domains, as in the classic Synthia~\cite{synthia} $\rightarrow$ CityScapes~\cite{cityscapes} setting. Thus, when copying objects from an image onto the other they are likely to end up in a reasonable context. This is not true for aerial images, where there is no consistent semantic structure (see \cref{fig:teaser} left).

To solve both problems, we propose a new mixing strategy for aerial segmentation across domains called \textbf{Hierarchical Instance Mixing} ({\ourMix}). {\ourMix} extracts from each semantic mask a set of connected components, akin to instance labels. The intuition is that aerial tiles often present very large stretches of land, divided into instances (e.g., forested areas separated by a road).
{\ourMix}  randomly selects from the individual instances a set of layers that will compose the binary mixing mask. This helps to mitigate the pixel imbalance between source and target domains in the artificial image.
Afterwards, {\ourMix} composes these sampled layers by sorting them based on the observation that there is a semantic hierarchy in the aerial scenes (\eg, cars lie on the road and roads lie on stretches of land). We use the pixel count of the instances to determine their order in this hierarchy, placing smaller layers on top of larger ones. 
While not optimal in some contexts (\eg, buildings should not appear on top of water bodies), this ordering also reduces the bias towards those categories with larger surfaces in terms of pixels as they are placed below the other layers of the mask (see \cref{fig:teaser} right).

Besides the mixing strategy itself, there is also the general problem that the effectiveness of the domain mixing is strongly dependent on the accuracy of the pseudo-labels generated on the target images during training.
This is especially true when the combination itself requires layering individual entities from either domain into a more coherent label.
A key factor for an effective domain adaptation using self-training is in fact the ability to produce consistent predictions, resilient to visual changes.
For this reason, we propose as a second contribution a \textbf{twin-head UDA architecture} in which two separate segmentation heads are fed with contrastive variations of the same images to improve pseudo-label confidence and make the model more robust and less susceptible to perturbations across domains, inevitably driving the model towards augmentation-consistent representations.

We test our complete framework on the LoveDA benchmark \cite{wang2021loveda}, the only dataset designed for evaluating unsupervised domain-adaption in aerial segmentation, where we exceed the current state-of-the-art. 
We further provide a comprehensive ablation study to assess the impact of the proposed solutions. The code will be made available to the public to foster the research in this field.

\section{\vspace{2pt}Related Work}
\begin{figure*}[t]
    \centering
    \includegraphics[width=\textwidth]{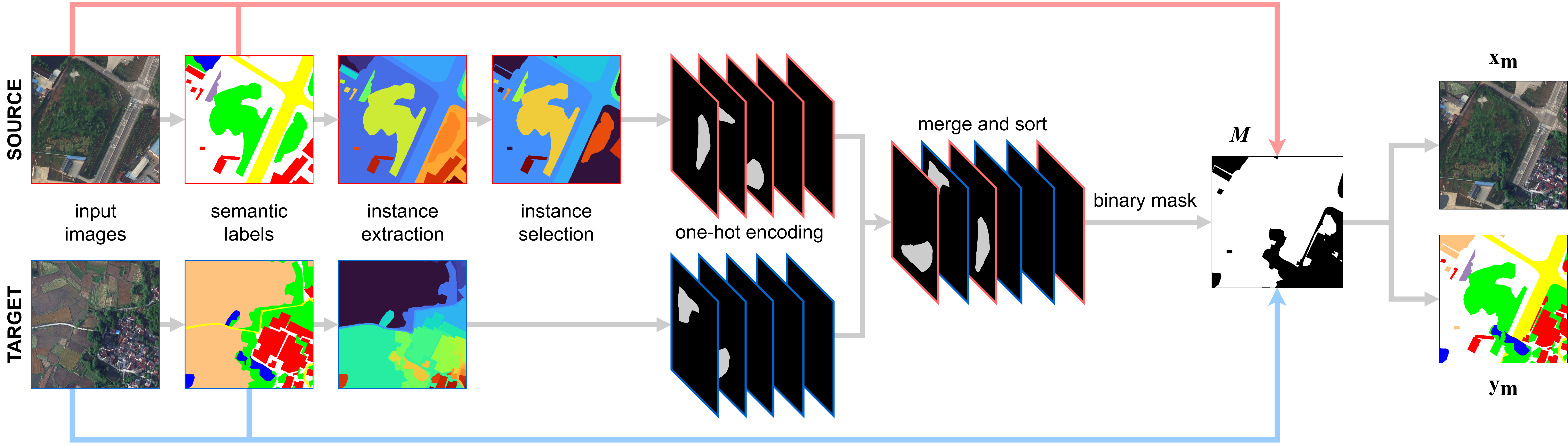}
    \vspace{0.2mm}
    \caption{HIMix operates by (i) extracting the connected components from the source label and target pseudo-label, (ii) selecting uniformly which instances should be mixed from $S$, (iii) merging source and target instances hierarchically based on instance size (smaller ones on top), and (iv) producing a binary mask $M$ to construct the final blended image $x_m$ and its label $y_m$.}
    \label{fig:id_mix}
\end{figure*}

\subsection{Aerial Semantic Segmentation}
Current semantic segmentation methods mostly rely on convolutional encoder-decoder architectures \cite{ref_fcn,ref_deeplab,ref_pspnet,ronneberger2015unet}, but the recent breakthroughs of vision Transformers introduced new effective encoder architectures such as ViT \cite{dosovitskiy2020vit}, Swin \cite{liu2021swin} or Twins \cite{chu2021twins}, as well as end-to-end segmentation approaches such as Segmenter \cite{strudel2021segmenter} and SegFormer \cite{xie2021segformer}.
Concerning the application to aerial images, despite the comparable processing pipeline as in other settings, there are peculiar challenges that demand for specific solutions. Firstly, aerial and satellite data often include multiple spectra besides the visible bands, which can be leveraged in different ways, such as including them as extra channels \cite{chiu2020agri_multispectral} or adopting multi-modal encoders \cite{ref_beyond_rgb}.
Visual features represent another major difference: unlike other settings, aerial scenes often display a large number of entities on complex backgrounds, with wider spatial relationships. In this case, attention layers \cite{niu2021rel_hybrid} or relation networks \cite{mou2019rel_relation} are employed to better model long-distance similarities among pixels. Another distinctive trait of aerial imagery is the top-down point of view and the lack of reference points that can be observed in natural images (e.g., sky always on top). This can be exploited to produce rotation-invariant features using ad-hoc networks \cite{han2021invariance, tavera2022aias}, or through regularization \cite{arnaudo2021invariance}.
Lastly, aerial images are characterized by disparities in class distributions, since these include small objects (e.g. cars) and large stretches of land. This pixel imbalance can be addressed with sampling and class weighting \cite{daformer}, or ad-hoc loss functions \cite{kevardec2019loss}.

\subsection{Domain Adaptation}
Domain Adaptation (DA) is the task of attempting to train a model on one domain while adapting to another. The main objective of domain adaptation is to close the \textit{domain shift} between these two dissimilar distributions, which are commonly referred to as the source and target domains.
The initial DA techniques proposed in the literature attempt to minimize a measure of divergence across domains by utilizing a distance measure such as the MMD~\cite{geng2011daml, pmlr-v37-long15, tzeng_mcd}.
Another popular approach to DA in Semantic Segmentation is adversarial training \cite{adaptsegnet, clan, fada, Tavera_2022_WACV}, which involves playing a min-max game between the segmentation network and a discriminator. This latter is responsible for discriminating between domains, whereas the segmentation network attempts to trick it by making features of the two distributions identical.
Other approaches, such as \cite{hoffman18cycada, wu2018dcan, yang2020fda}, employ image-to-image translation algorithms to generate target pictures styled as source images or vice versa, while \cite{transnorm} discovers the major bottleneck with domain adaptation in the batch normalization layer.
More recent methods like \cite{pycda, cbst, iast} use self-learning techniques to generate fine pseudo-labels on target data to fine-tune the model, whereas \cite{dacs, daformer} combine self-training with class mix to reduce low-quality pseudo-labels caused by domain shifts among the different distributions. 

These mixing algorithms are very effective on data with a consistent semantic organization of the scene, such as in self-driving scenes \cite{cityscapes, idda}. In these scenarios, naively copying half of the source image onto the target image increases the likelihood that the semantic elements will end up in a reasonable context. This is not the case with aerial imagery (see \cref{fig:teaser}).
{\ourMix} not only mitigates this problem, but it also reduces the bias towards categories with larger surfaces.

\section{Method}
\subsection{Problem statement}
We investigate the aerial semantic segmentation task in the context of unsupervised domain adaption (UDA). Let us define as $\mathcal{X}$ the set of RGB images constituted by the set of pixels $\mathcal{I}$, and as $\mathcal{Y}$ the set of semantic masks associating a class from the set of semantic classes $\mathcal{C}$ to each pixel $i \in \mathcal{I}$. 
We have two sets of data accessible at training time: (i) a set of annotated images from the source domain, denoted as $X_{s} = \{(x_{s}, y_{s})\}$ with $x_{s}\in \mathcal{X}$ and $y_{s} \in \mathcal{Y}$; (ii) a set of $N_{t}$ unlabelled images from the \textit{target} domain, denoted as $X_{t} = \{(x_{t})\}$ with $x_{t}\in \mathcal{X}$.

The goal is to find a parametric function $f_\theta$ that maps a RGB image to a pixel-wise probability, \ie, $f_\theta: \mathcal{X} \rightarrow \mathbb{R}^{|\mathcal{I}|\times|\mathcal{C}|}$, and evaluate it on unseen images from the target domain. In the following, we indicate the model output in a pixel $i$ for the class c as $p_i^c$, \ie, $p_i^c(x) = f_\theta(x)[i,c]$.
The parameters $\theta$ are tuned to minimize a categorical cross-entropy loss defined as

\begin{equation}
    L_{\text{seg}}(x, y) = - \frac{1}{|\mathcal{I}|} \sum_{i \in \mathcal{I}} \sum_{c \in \mathcal{C}}  y_i^c \log(p_i^c(x)),
    \label{eq:xe}
\end{equation}
where $y_i^c$ represents the ground truth annotation for the pixel $i$ and class $c$.

\subsection{Framework}
We present an end-to-end trainable UDA framework based on the use of target pseudo-labels. To better align domains, we construct artificial images using our {\ourMix} strategy (\ref{sec:idmix}), which generates mixed images exploiting the instances produced both from the source ground truth and the target pseudo-label.
Rather than using a secondary teacher network derived from the student as an exponential moving average as in \cite{dacs, daformer}, we propose a twin-head architecture (\ref{sec:twinhead}) with two separate decoders trained in a contrastive fashion to provide finer target pseudo labels.

\subsection{Hierarchical Instance Mixing} \label{sec:idmix}
Given the pairs $(x_s, y_s)$ and $(x_t, \hat{y}_t)$, where $\hat{y}_t = f_\theta(x_t)$ are the pseudo-labels computed from the model prediction on the target domain, the purpose of the mixing strategy is to obtain a third pair, namely $(x_m, y_m)$, whose content is derived from both source and target domains using a binary mask $M$. 
While techniques based on ClassMix have been successfully applied in many UDA settings, we discover that the same may not be optimal in the aerial scenario since it superimposes parts of the source domain onto the target without taking into consideration of their semantic hierarchy (\eg, cars appear on top of roads, not vice versa). 
In contrast, we propose an Hierarchical Instance Mixing strategy ({\ourMix}), which is composed of two subsequent steps: (i) \textit{instance extraction} and (ii) \textit{hierarchical mixing}.

\myparagraph{Instance extraction.}
Aerial tiles often present uniform land cover features, with many instances of the same categories in the single image. In the absence of actual instance labels, this peculiarity can be exploited to separate semantic annotations into connected components. 
Here a connected component is a set of pixels that have the same semantic label and such that for any two pixels in this set there is a path between them that is entirely contained in the same set.
\Cref{fig:id_mix} illustrates an example of this process, with a forest that is separated in two instances by a road.
This increases the number of regions which can be randomly selected for the mixing phase, thus mitigating the pixel unbalance in the final mixed sample between source and target domains.
Note that this procedure is applied on the concatenation of source and target label.

\begin{figure*}[t]
    \centering
    \includegraphics[width=1.0\textwidth]{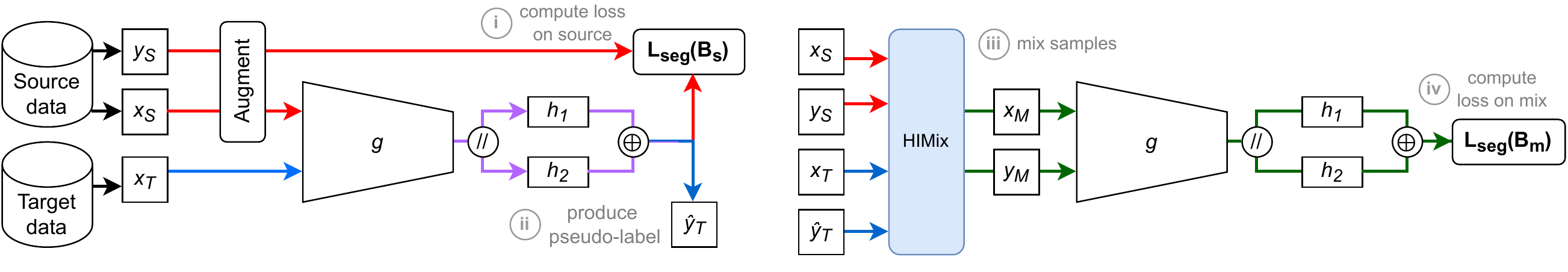}
    \vspace{0.2mm}
    \caption{Our framework training: (i) standard training is carried out on source, (ii) pseudolabels are generated on target through majority voting between each head output, (iii) source and target samples are mixed together and (iv) segmentation loss is computed on mixed pairs.}
    \label{fig:architecture}
\end{figure*}

\myparagraph{Hierarchical mixing.} 
We observe that instances in aerial imagery have an inherent hierarchy that is dictated by their semantic categories. In other words, land cover categories such as \textit{barren} or \textit{agricultural} frequently appear in the background w.r.t. smaller instances such as \textit{roads} or \textit{buildings}.
The mixing step follows this hierarchy when combining the instances from source and target, and it is illustrated in \cref{fig:id_mix}.
First, both sets of instance labels are encoded into a one-hot representation, so that each component yields its own mask layer. 
Then both stacks of layers are merged together and sorted by their pixel count, with the larger layers on the bottom. Finally, a reduction from top to bottom projects the 3D tensor into a 2D binary mask $M$, where positive values indicate \textit{source} pixels, and null values indicate \textit{target} pixels.

\subsection{Twin-Head Architecture} \label{sec:twinhead}
State-of-the-art, self-training UDA strategies, such as \cite{daformer}, make use of \textit{teacher-student} networks to improve the consistency of the pseudo-labels. Albeit dealing with consistency in time, teacher-based approaches do not directly cope with geometric or stylistic consistency.
We propose a twin-head segmentation framework to directly address this, providing more consistent pseudo-labels and outperforming the standard tested methodologies, as shown in the ablation study \ref{sec:ablation}.
Our architecture (see Fig. \ref{fig:architecture}) comprises a shared encoder $g$, followed by two parallel and lightweight segmentation decoders, $h_1$ and $h_2$. Training is carried out end to end, exploiting annotated source data and computing pseudo-labels from target images online, as detailed hereinafter.

\myparagraph{Source training.} With the purpose of driving the model towards augmentation-consistent representations, we feed the two heads with variations of the same image in a contrastive fashion. More specifically, given a source image $x_s$ we alter it with a sequence of random geometric (\textit{horizontal flipping}, \textit{rotation}) and photometric augmentations (\textit{color jitter}), obtaining new pairs of samples.
Specifically, at each iteration, the final input is composed of $B_s = (x_s \mathbin\Vert \tilde{x}_s, y_s \mathbin\Vert \tilde{y}_s)$,
where $x_s$ and $y_s$ represent the original batch of images and respective annotations, while $\tilde{x}_s$ and $\tilde{y}_s$ represent the same samples, altered by the geometric and photometric transformations, \ie, $\tilde{x}_s = T_p(T_g(x_s))$, and $\tilde{y}_s = T_g(y_s)$.
The full augmented batch $B_s$ is first forwarded to the shared encoder module $g$, producing a set of features. The latter, containing information derived from the images and its augmented variants, are split and forwarded to the two parallel heads, effectively obtaining two comparable outputs, $h_1(g(x_s))$ and $h_2(g(\tilde{x}_s))$.
A standard cross-entropy loss, as shown in Eq. \ref{eq:xe}, is computed on both segmentation outputs.
Working independently on different variations of the same images, the two heads can evolve in different ways while trying to minimize the same objective function. 
Using the same encoder yields a more robust, contrastive-like feature extraction that is less susceptible to perturbations. This is essential for producing more stable and precise pseudo-labels.

\myparagraph{Mix training.} The twin-head architecture is expressly designed to generate more refined pseudo-labels. Given an unlabeled target image $x_t$, the probabilities after forwarding the image to both heads $\sigma(h_1(g(x_t)))$ and $\sigma \left( h_2(g(\tilde{x}_t)) \right)$ are compared, where $\sigma$ indicates the softmax function. In order to extract a single pseudo-label, the most confident output is selected for each pixel. Formally, for each position $i$ the output score is computed as $p_i^c = max(\sigma_i(h_1(g(x_t)), \sigma_i \left( h_2(g(\tilde{x}_t)) \right))$, selecting the maximum value between the two.
Once $p_i^c$ is derived, the pseudo-label $\hat{y}_t$ necessary for class-mix is generated through:

\begin{equation}
    \hat{y}_t^{(i,c)} = [c = argmax_c p_i^c (x_t)].
    \label{eq:pl}
\end{equation}

At this point, the mixed pairs of inputs can be computed through {\ourMix}, as described in previous sections, obtaining $(x_m, y_m)$ as a composition of the source and target samples.
Similar to source training, an augmented batch $B_m = (x_m \mathbin\Vert \tilde{x}_m, y_m \mathbin\Vert \tilde{y}_m)$ is computed through geometric and photometric transformations, then fed to the model to compute $L_{seg}(B_m)$.
To reduce the impact of low-confidence areas, a pixel-wise weight map $w_m$ is generated. Similar to \cite{dacs, daformer}, the latter is computed as percentage of valid points above threshold. Formally, for each pixel $i$:

\begin{equation}
    w_m^i =
    \begin{cases}
    1, & i \in y_s \\
    \dfrac{m_{\tau}}{\left|\mathcal{I}\right|}, & i \in \hat{y}_t \\
    \end{cases}
    \label{eq:weight_map}
\end{equation}
where $m_{\tau}$ represents the Max Probability Threshold \cite{li2019bidirectional} computed over pixels belonging to the pseudo-label as follows:

\begin{equation}
   m_{\tau}^i = \mathds{1}_{[argmax_c p_i^c (x_t) > \tau]}, 
    \label{eq:pl_th}
\end{equation}

In practice, each pixel of the mixed label is either weighted as $1$ for regions derived from the source domain, or by a factor obtained as the  number of pixels above the confidence threshold, normalized by the total amount of pixels. 
Note that during all of these computations the gradients are not propagated. The training procedure is  detailed in \cref{pc:pseudo_code}.

\begin{algorithm}[!ht]
\SetAlgoLined
\SetKwBlock{Source}{Train on \textit{source} $\mathcal{X}_s$}{end}
\SetKwBlock{Mix}{Mix \textit{source} and \textit{target} pairs}{end}
\SetKwBlock{Target}{Train on \textit{mixed} $\mathcal{X}_m$ pairs}{end}

\textbf{Initialize:} 
    \\Model $f_\theta: \mathcal{X} \rightarrow \mathbf{R}^{|\mathcal{I}| \times |\mathcal{Y}|}$ with encoder $g$ and twin heads $h_1, h_2$\;
\textbf{Input:} 
    $\mathcal{X}_s$ source domain with $N_s$ pairs $(x_s, y_s)$, $x_s  \in \mathcal{X}, y_s \in \mathcal{Y}$ and semantic classes $\mathcal{C}$\;
    $\mathcal{X}_t$ target domain with $N_t$ images $x_t$, lacking ground truth labels\;
\textbf{Output:}
    $y=\{\text{argmax}_{c\in \mathcal{Y}} p_i^c\}^N_{i=1}$, where $p_i^c$ the model prediction of pixel $i$ for class $c$ and $\mathcal{Y}$ the label space\;
 
\While{epoch in max\_epochs}{
    \While{$x_s, y_s, x_t$ in $\mathcal{X}_s \times \mathcal{X}_t$}{
        \Source{
            Compute augmented source batch $B_s = (x_s \mathbin\Vert \tilde{x}_s, y_s \mathbin\Vert \tilde{y}_s)$\;
            Train $f_\theta$ on source labels with $L_{seg}(B_s)$\;
        }
        \Mix{
             Compute pseudo-labels via majority voting $\hat{y}_t = max\left(h_1(g(x_t)), T_g^{-1}(h_2(g( \tilde{x}_t)))\right)$\;
             Extract source instance labels $i_s = CCL(y_s)$ with instances $\in K_s$\;
             Extract target instance pseudo-labels $i_t = CCL(\hat{y}_t)$ with instances $\in K_t$\;
             Compute one-hot encoded labels, sorted by pixel size as $1_{m} = sorted\left(1_{K_s}(i_s) \mathbin\Vert 1_{K_t}(i_t) \right)$\;
             Reduce \textit{z} axis to 2D indexed mask $m = argmax_z 1_{m}(i, j, z)$\;
             Binarize mask $\forall i,j \in m, \ M =
            \begin{cases}
                  1 & if \ m(i, j) \in K_s  \\
                  0 & if \ m(i, j) \in K_t 
             \end{cases}
            $\;
            Compute mixed image and labels as \;
            $x_m = M \odot x_s + (1 - M) \odot x_t$\;
            $y_m = M \odot y_s + (1 - M) \odot \hat{y}_t$\;
            Compute $w_m$ as in Eq. \ref{eq:weight_map}
        }
        \Target{
            Compute augmented mixed batch $B_m = (x_m \mathbin\Vert \tilde{x}_m, y_m \mathbin\Vert \tilde{y}_m)$\;
            Train $f_\theta$ on mixed samples with $L_{seg}(B_m)$, weighted by $w_m$\;
        }
    }
}
\caption{Twin-head UDA training procedure}
\label{pc:pseudo_code}
\end{algorithm}

\section{Experiments}
\subsection{Training Details}
We assess the performance of our approach on the LoveDA dataset \cite{wang2021loveda}. According to that benchmark, we conduct two series of unsupervised domain adaptation experiments: \textit{rural}$\to$\textit{urban} and \textit{urban}$\to$\textit{rural}. We measure the performance on the \textit{test set} of each target domain. 

\myparagraph{Dataset.} 
To our knowledge, the LoveDA dataset \cite{wang2021loveda} is the only open and free collection of land cover semantic segmentation images in remote sensing explicitly designed for UDA. Both urban and rural areas are included in the training, validation, and test sets. Data is gathered from 18 different administrative districts in China. The urban training set has 1156 images, while the rural training set contains 1366 images. Each image is supplied in a tiled format of 1024x1024 pixels annotated with seven categories.

\myparagraph{Metric.}
Following \cite{wang2021loveda} we use the averaged Intersection over Union (mIoU) metric to measure the accuracy of all the experiments conducted.

\myparagraph{Baselines.} 
Our method is compared to various cutting-edge UDA methods. The first baseline we consider is the Source Only model, which is a network that has only been trained using the source dataset. We look at MMD's \cite{tzeng_mcd} original metric-based methodology. Then, we compare two alternative UDA approaches: the adversarial training strategy, with  AdaptSegNet\cite{adaptsegnet}, FADA\cite{fada}, CLAN\cite{clan}, and TransNorm\cite{transnorm}, and the self-training technique, with CBST\cite{cbst}, PyCDA\cite{pycda}, IAST\cite{iast}, DACS\cite{dacs} and DAFormer \cite{daformer}.

\myparagraph{Implementation.}
To implement our solution we leverage the \textit{mmsegmentation} framework, that is based on PyTorch. We train each experiment on a NVIDIA Titan GPU with 24GB of RAM. We refer to DAFormer \cite{daformer} for the architecture and configuration of hyperparameters. We use the MiT-B5 model \cite{xie2021segformer} pretrained on ImageNet as the encoder of our method while the segmentation decoder module corresponds to the SegFormer head \cite{xie2021segformer}. We train on every setting for 40k iterations using AdamW as optimizer. The learning rate is set to $6x10^{-5}$, weight decay of $0.01$, betas to $(0.9, 0.99)$. We also adopt a polynomial decay with a factor of $1.0$ and warm-up for 1500 iterations. To cope with possible variations, every experiment presented has been obtained as the average over three seeds $\{0,1,2\}$. Training is performed on random crops, by augmenting data through random resizing in the range $[0.5, 2.0]$, horizontal and vertical flipping, and rotation of 90 degrees with probability $p=0.5$, together with random photometric distortions (i.e., brightness, saturation, contrast and hue). As \cite{dacs, daformer}, we set $\tau=0.968$. The final inference on the test set is instead performed on raw images without further transformations.

\subsection{Results}
\begin{table}[t]
\centering
\begin{adjustbox}{width=1.0\columnwidth}

\begin{tabular}{lrrrrrrrr} 
\midrule
\multicolumn{1}{c}{\textbf{Method}} & \multicolumn{1}{l}{\begin{sideways}\textbf{Backg.}\end{sideways}} & \multicolumn{1}{l}{\begin{sideways}\textbf{Building}\end{sideways}} & \multicolumn{1}{l}{\begin{sideways}\textbf{Road}\end{sideways}} & \multicolumn{1}{l}{\begin{sideways}\textbf{Water}\end{sideways}} & \multicolumn{1}{l}{\begin{sideways}\textbf{Barren}\end{sideways}} & \multicolumn{1}{l}{\begin{sideways}\textbf{Forest}\end{sideways}} & \multicolumn{1}{l}{\begin{sideways}\textbf{Agric.}\end{sideways}} & \multicolumn{1}{c}{\textbf{mIoU}} \\ 
\midrule
\rowcolor{lightgray} Source only & 24.2 & 37.0 & 32.6 & 49.4 & 14.0 & 29.3 & 35.7 & 31.7 \\
MCD \cite{tzeng_mcd} & 25.6 & 44.3 & 31.3 & 44.8 & 13.7 & 33.8 & 26.0 & 31.4 \\
AdaptSeg \cite{adaptsegnet} & 26.9 & 40.5 & 30.7 & 50.1 & 17.1 & 32.5 & 28.3 & 32.3 \\
FADA \cite{fada} & 24.4 & 33.0 & 25.6 & 47.6 & 15.3 & 34.4 & 20.3 & 28.7 \\
CLAN \cite{clan} & 22.9 & 44.8 & 26.0 & 46.8 & 10.5 & 37.2 & 24.5 & 30.4 \\
TransNorm \cite{transnorm} & 19.4 & 36.3 & 22.0 & 36.7 & 14.0 & \underline{40.6} & 03.3 & 24.6 \\
PyCDA \cite{pycda} & 12.4 & 38.1 & 20.5 & 57.2 & \underline{18.3} & 36.7 & 41.9 & 32.1 \\
CBST \cite{cbst} & 25.1 & 44.0 & 23.8 & 50.5 & 08.3 & 39.7 & 49.7 & 34.4 \\
IAST \cite{iast} & 30.0 & 49.5 & 28.3 & 64.5 & 02.1 & 33.4 & 61.4 & 38.4 \\
DACS \cite{dacs} & 20.1 & 50.5 & 35.9 & 60.6 & 09.9 & 35.4 & 17.5 & 32.9 \\
DAFormer \cite{daformer} & 29.5 & 57.9 & 41.8 & 67.1 & 07.6 & 35.3 & 48.1 & 41.0 \\ 
\midrule
\textbf{Ours} & \underline{31.5}  & \underline{59.6} & \underline{51.5} & \underline{68.1} & 08.2 & 37.4 & \underline{53.9} & \textbf{44.3}
\end{tabular}

\end{adjustbox}
\caption{Urban$\to$Rural experiments.}
\label{table:u2r_exps}
\end{table}
\begin{table}[t]
\centering
\begin{adjustbox}{width=1.0\columnwidth}

\begin{tabular}{lrrrrrrrr} 
\midrule
\multicolumn{1}{c}{\textbf{Method}} & \multicolumn{1}{l}{\begin{sideways}\textbf{Backg.}\end{sideways}} & \multicolumn{1}{l}{\begin{sideways}\textbf{Building}\end{sideways}} & \multicolumn{1}{l}{\begin{sideways}\textbf{Road}\end{sideways}} & \multicolumn{1}{l}{\begin{sideways}\textbf{Water}\end{sideways}} & \multicolumn{1}{l}{\begin{sideways}\textbf{Barren}\end{sideways}} & \multicolumn{1}{l}{\begin{sideways}\textbf{Forest}\end{sideways}} & \multicolumn{1}{l}{\begin{sideways}\textbf{Agric.}\end{sideways}} & \multicolumn{1}{c}{\textbf{mIoU}} \\ 
\midrule
\rowcolor{lightgray} Source only & 43.3 & 25.6 & 12.7 & 76.2 & 12.5 & 23.3 & 25.1 & 31.3 \\
MCD \cite{tzeng_mcd} & 43.6 & 15.4 & 12.0 & 79.1 & 14.3 & 33.1 & 23.5 & 31.5 \\
AdaptSeg \cite{adaptsegnet} & 42.4 & 23.7 & 15.6 & 82.0 & 13.6 & 28.7 & 22.1 & 32.6 \\
FADA \cite{fada} & 43.9 & 12.6 & 12.8 & 80.4 & 12.7 & 32.8 & 24.8 & 31.4 \\
CLAN \cite{clan} & 43.4 & 25.4 & 13.8 & 79.3 & 13.7 & 30.4 & 25.8 & 33.1 \\
TransNorm \cite{transnorm} & 33.4 & 05.0 & 03.8 & 80.8 & 14.2 & 34.0 & 17.9 & 27.7 \\
PyCDA \cite{pycda} & 38.0 & 35.9 & 45.5 & 74.9 & 07.7 & 40.4 & 11.4 & 36.3 \\
CBST \cite{cbst} & 48.4 & 46.1 & 35.8 & 80.1 & 19.2 & 29.7 & 30.1 & 41.3 \\
IAST \cite{iast} & 48.6 & 31.5 & 28.7 & 86.0 & \underline{20.3} & 31.8 & 36.5 & 40.5 \\
DACS \cite{dacs} & 46.0 & 31.6 & 33.8 & 76.4 & 16.4 & 29.3 & 27.7 & 37.3 \\
DaFormer \cite{daformer} & 49.2 & 47.7 & 55.2 & \underline{86.6} & 16.5 & 39.5 & 30.8 & 46.5 \\ 
\midrule
\textbf{Ours} & \underline{49.3} & \underline{55.0} & \underline{55.4} & 86.0 & 17.1 & \underline{41.2} & \underline{36.9} & \textbf{48.7}
\end{tabular}

\end{adjustbox}
\caption{Rural$\to$Urban experiments.}
\label{table:r2u_exps}
\end{table}

\begin{table}[t]
\centering

\resizebox{1.0\columnwidth}{!}{
\begin{tabular}{c|c|c|c|c|c|c}
ID & Twin & Class & Instance & Hierarc. & mIoU & mIoU \\ 
& Head & Mix & Mix & Mix & U2R & R2U  \\
\midrule
1 &  & \cmark & & & 41.0 $\pm$ 0.33 & 46.5 $\pm$ 0.41 \\ 
2 & &  & \cmark & \cmark & 43.4 $\pm$ 0.76 & 47.6 $\pm$ 0.10 \\
3 & \cmark & \cmark & & & 42.9 $\pm$ 0.35 & 47.1 $\pm$ 0.34 \\
4 & \cmark &  & \cmark &   & 43.2 $\pm$ 0.35 & 47.4 $\pm$ 0.16 \\
5 & \cmark &  & \cmark & \cmark & \textbf{44.3} $\pm$ 0.39 &  \textbf{48.7} $\pm$ 0.06\\

\end{tabular}
}

\caption{Ablation study on our twin-head architecture and {\ourMix} strategy.}
\label{table:ablation}
\end{table}

\begin{figure*}[t]
    \centering
    \includegraphics[width=0.95\textwidth]{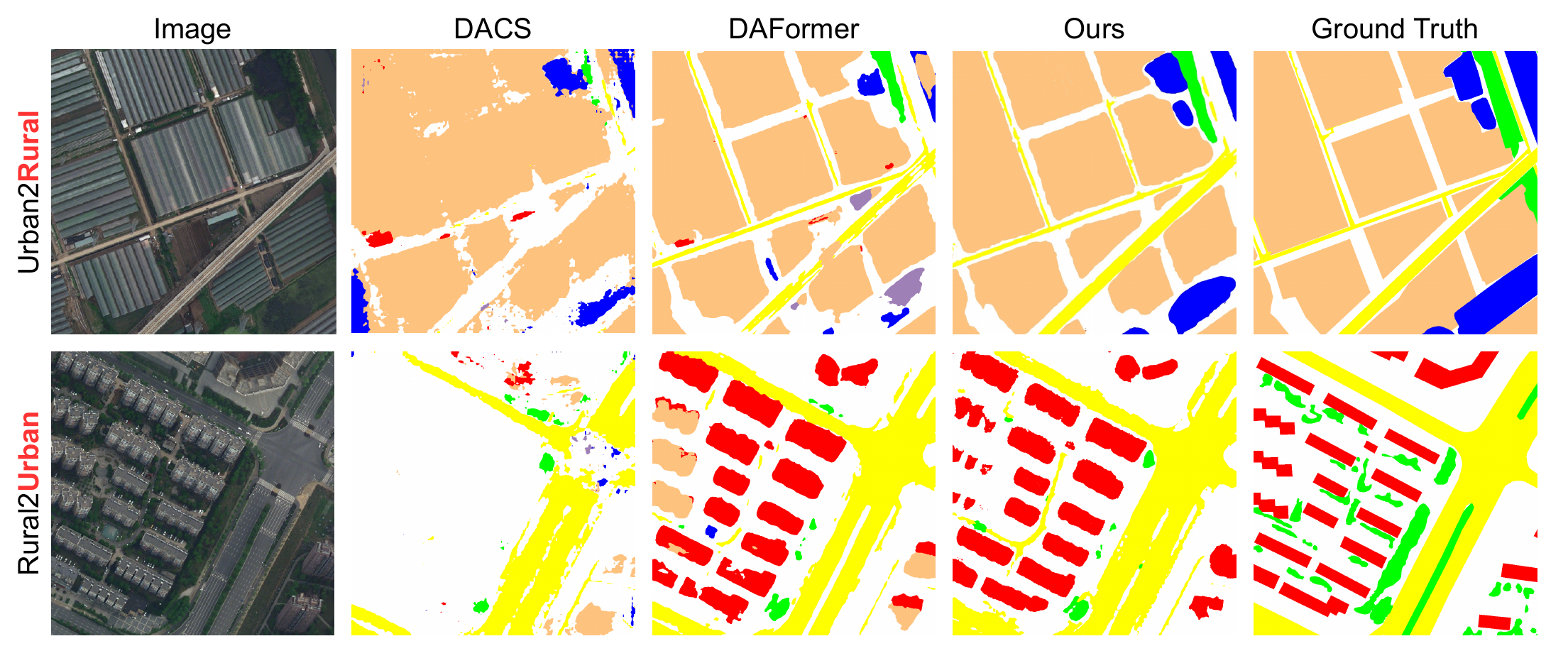}
    \caption{Qualitative results in the two settings \textit{Urban$\to$Rural} and \textit{Rural$\to$Urban} after testing on \textbf{\color{red}{target}} domain.}
    \label{fig:qualitatives}
\end{figure*}

\myparagraph{Urban$\to$Rural.}
The results for this set of experiments are reported in \cref{table:u2r_exps}. They corroborate the complexity of the task due to a strong and inconsistent class distribution in the source domain, which is dominated by urban scenes with a mix of buildings and highways but few natural items. This causes a negative transfer to the target domain, since both adversarial strategies and self-training procedures achieve overall performance equivalent to, if not worse than, the Source Only model. Specifically, when we evaluate the best performing Adversarial Training technique, which is represented by CLAN, we gain just a $+1.8$ improvement over the Source Only model. Self-training approaches have shown to be the most effective. DACS, which introduces the class mix strategy, improves the \textit{Source Only} model by $+1.2$, while DAFormer, which uses a Transformer backbone and the same class mix strategy as DACS, outperforms the \textit{Source Only} model by $+9.3$. Our approach, which combines both the twin-head architecture and the innovative class mix, outperforms the \textit{Source Only} model by the wide margin of $+12.6$ and it exceeds its closest competitor (DAFormer) by $+3.3$. {\ourMix} exhibits its ability to boost rural and underrepresented classes, such as \textit{agriculture}, as also evidenced by qualitative results in \cref{fig:qualitatives}. In comparison to DACS and DAFormer, our technique recognizes and classifies better contours and classes, such as \textit{water}, despite their underrepresentation in the source domain. This is also true in common categories with different visual features such as \textit{road}, which can appear in paved and unpaved variants.

\myparagraph{Rural$\to$Urban.}
The results for this set of experiments are summarized on \cref{table:r2u_exps}. The source domain in this scenario is dominated by large-scale natural objects and a few manmade samples. Nonetheless, the models under consideration are capable of effectively transferring the knowledge even in these underrepresented categories. Self-learning approaches outperform adversarial methods, getting an averaged boost of $+9.1$ over the Source Only model, whereas adversarial training methods achieve comparable accuracies. In terms of mIoU, the two most performant self-training models and our closest competitor surpass the Source Only model by $+6.0$ and $+15.2$, respectively.
In comparison, our strategy gains a $+17.4$ boost over the Source Only model, outperforming DACS and DAFormer by $+11.4$ and $+2.2$, respectively.
In this case, the qualitative results in \cref{fig:qualitatives} support the superior ability of our model to discern between rural and urban classes. While DACS does not recognize \textit{buildings} and DAFormer misclassifies parts of them as \textit{agricultural} terrain, our model demonstrates its efficacy in minimizing the bias towards those categories with larger surfaces providing results close to the ground truth.

\begin{figure}[h!t]
    \centering
    \includegraphics[width=0.9\columnwidth]{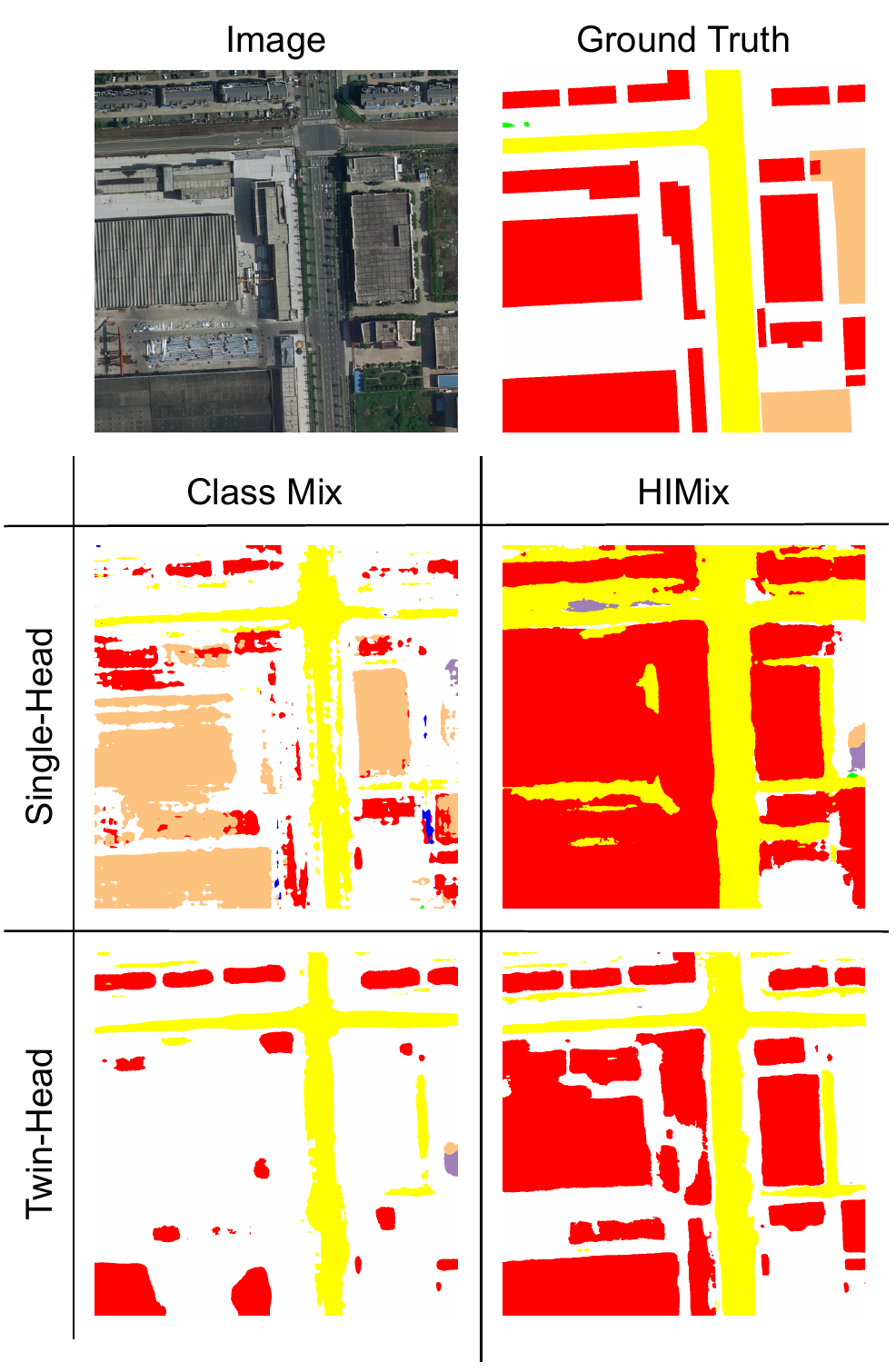}
    \caption{Qualitative comparison of Single or Twin-Head architectures using Standard Class Mix or our {\ourMix}.}
    \label{fig:ablation}
\end{figure}

\subsection{Ablation}\label{sec:ablation}

\myparagraph{Twin-Head and {\ourMix}.}
To demonstrate the effectiveness of the twin-head architecture, we compare it to the traditional single-head structure, which generates pseudo-labels using a secondary teacher network derived from the student as an exponential moving average. This study also demonstrates the potential of the {\ourMix} when paired with traditional single-head training. For both the settings, we perform an extensive ablation study considering the MiT-B5 \cite{xie2021segformer} as the backbone and we report the results in \cref{table:ablation}.

The twin-head design paired with the Standard Class Mix (line 3) is more performing than the single-head architecture (line 1), implying that our solution is better at providing finer pseudo-labels with correct class segmentation, as also shown in the first column of \cref{fig:ablation}.

{\ourMix} increases recognition performance even when paired with a single-head architecture (line 2), particularly for categories with a lower surface area in terms of pixels, which are placed below those with larger surfaces when using the Standard Class Mix. That is why, in the top-left image of \cref{fig:ablation}, the model is unable to grasp their semantics effectively and erroneously classifies \textit{building} as \textit{agricultural} pattern. In comparison, {\ourMix} can accurately distinguish \textit{buildings} (top-right picture in \cref{fig:ablation}) even though the prediction has poorly defined contours.

The best results are obtained when the twin-head ability to provide an enhanced segmentation map is combined with the {\ourMix} ability to maintain a correct semantic structure (line 5), yielding the best results in terms of accuracy and finer segmentation map, as shown in the bottom-right image of \cref{fig:ablation}.  

We finally ablate the different components of our {\ourMix} to assess each term's contribution to overall performance (lines 4-5). The Hierarchical Mixing always increases the Instance Extraction by $+1.1$ and $+1.3$ in the two Urban$\to$Rural and Rural$\to$Urban scenarios, respectively.

\section{Conclusions}
We investigated the problem of Unsupervised Domain Adaptation (UDA) in aerial Semantic Segmentation, showing that the peculiarities of aerial imagery, principally the lack of structural consistency and a significant disparity in semantic class extension, must be taken into consideration. We addressed these issues with two contributions. First, a novel domain mixing method that consists of two parts: an instance extraction that chooses the connected components from each semantic map and a hierarchical mixing that sorts and fuses the instances based on their pixel counts. Second, a twin-head architecture that produces finer pseudo labels for the target domain, improving the efficacy of the domain mixing. 
We demonstrated the effectiveness of our solution with a comprehensive set of experiments  on the LoveDA benchmark. 

\myparagraph{Limitations.}
Despite the excellent results, we observed that our solution has worse performance than the source only model in the \textit{barren} class, particularly in the Urban$\to$Rural scenario. This is possibly due to the large disparity in absolute pixels count between source and target domains in this category. Additionally, the twin-head architecture, while its superior performance, has a greater number of parameters that slow down the training (approximately 15h).

\myparagraph{Future Works.} We will evaluate lighter segmentation heads and other contrastive techniques to accelerate overall training and improve performance, particularly on specific semantic classes.










\bibliographystyle{IEEEtran.bst} 
\bibliography{IEEEabrv,bibfile}
\end{document}